\documentclass[10pt,twocolumn,letterpaper]{article}

\usepackage[pagenumbers]{cvpr}        

\usepackage{bibunits}

\definecolor{cvprblue}{rgb}{0.21,0.49,0.74}
\usepackage[pagebackref,breaklinks,colorlinks,allcolors=cvprblue]{hyperref}

\definecolor{daqfirstcolor}{HTML}{E5CCFF}
\definecolor{daqsecondcolor}{HTML}{FFE5CC}
\definecolor{daqthirdcolor}{HTML}{FFFFCC}

\DeclareRobustCommand{\daqhighlight}[2]{%
    \begingroup
    \setlength{\fboxsep}{0pt}%
    \colorbox{#1}{\color{black}\strut#2}%
    \endgroup
}

\newcommand{\daqfirst}[1]{\daqhighlight{daqfirstcolor}{#1}}
\newcommand{\daqsecond}[1]{\daqhighlight{daqsecondcolor}{#1}}
\newcommand{\daqthird}[1]{\daqhighlight{daqthirdcolor}{#1}}

\def\confName{CVPR}
\def\confYear{2027}

\title{DAQ: Decision-Aligned Post-Training Quantization for Spiking Transformers}

\author{
    Jialin Chen$^{1,\dagger}$,
    Xiaochen Zhao$^{1,\dagger}$,
    Aili Wang$^{1,2,3,*}$\\
    $^{1}$ ZJU-UIUC Institute, Zhejiang University\\
    $^{2}$ Brain Machine Intelligence Lab, Zhejiang University \\
    $^{3}$ College of Information Science and Electronic Engineering, Zhejiang University
}

\begin{document}

\maketitle

\begingroup
\makeatletter
\renewcommand{\thefootnote}{}
\renewcommand{\@makefntext}[1]{\noindent#1}
\begin{NoHyper}
\footnotetext{%
    \raggedright
    $^{\dagger}$ Equal contribution.\\
    $^{*}$ Corresponding author:
    \texttt{ailiwang@intl.zju.edu.cn}.
}
\end{NoHyper}
\makeatother
\endgroup

\begin{abstract}
Spiking transformers combine attention-based representation learning with sparse spike communication, yet their weights and membrane variables still incur substantial storage and computation overhead. Post-training quantization (PTQ) offers a practical way to reduce these overheads, but preserving neuronal firing under weight and membrane quantization remains challenging. In this paper, we propose Decision-Aligned Post-Training Quantization (DAQ), a method that calibrates the quantizers to align neuronal firing decisions with those of a floating-point reference. We analyze how errors from these quantizers affect firing at decision boundaries and how they propagate over time steps in recurrent neurons. Based on this analysis, we formulate an objective that encourages quantized membrane variables to remain within the reference firing regions along the quantized model's own state trajectory. Experiments across multiple spiking transformer architectures on image classification, semantic segmentation, and event-based tracking demonstrate the effectiveness of DAQ with 4-bit weights and membrane variables. On FE108, DAQ achieves 58.8\% AUC and 90.9\% PR with SDTrack, while reducing energy by 66.2\%.
\end{abstract}

\section{Introduction}
\label{sec:daq-introduction}

Spiking transformers combine attention-based representation
learning with discrete spike communication and have been
applied to image classification, semantic segmentation,
and event-based object
tracking~\cite{yao2023spike,yao2025scaling,lei2025spike2former,shan2025sdtrack}.
Their synaptic weights and membrane variables still require
substantial storage and processing
resources~\cite{hasssan2024spquant,wei2026tp}.
Post-training quantization (PTQ) reduces numerical precision
by calibrating pretrained models on a small dataset,
avoiding the cost of full
retraining~\cite{li2021brecq,wu2025fima}.

We study PTQ of both synaptic weights and membrane variables
in spiking transformers.
The effect of a perturbation on firing depends on its direction
and the potential's position relative to a decision
boundary~\cite{yao2023spike,yao2025scaling}.
For a potential just below the firing threshold, a positive
perturbation can trigger a spike, whereas a negative perturbation
of the same magnitude leaves the neuron silent.
Integer-valued neurons exhibit analogous changes at boundaries
between adjacent spike counts.
Preserving firing therefore requires the quantized potential
to remain in the region associated with the original output.

\begin{figure}[t]
    \centering
    \includegraphics[width=\columnwidth]{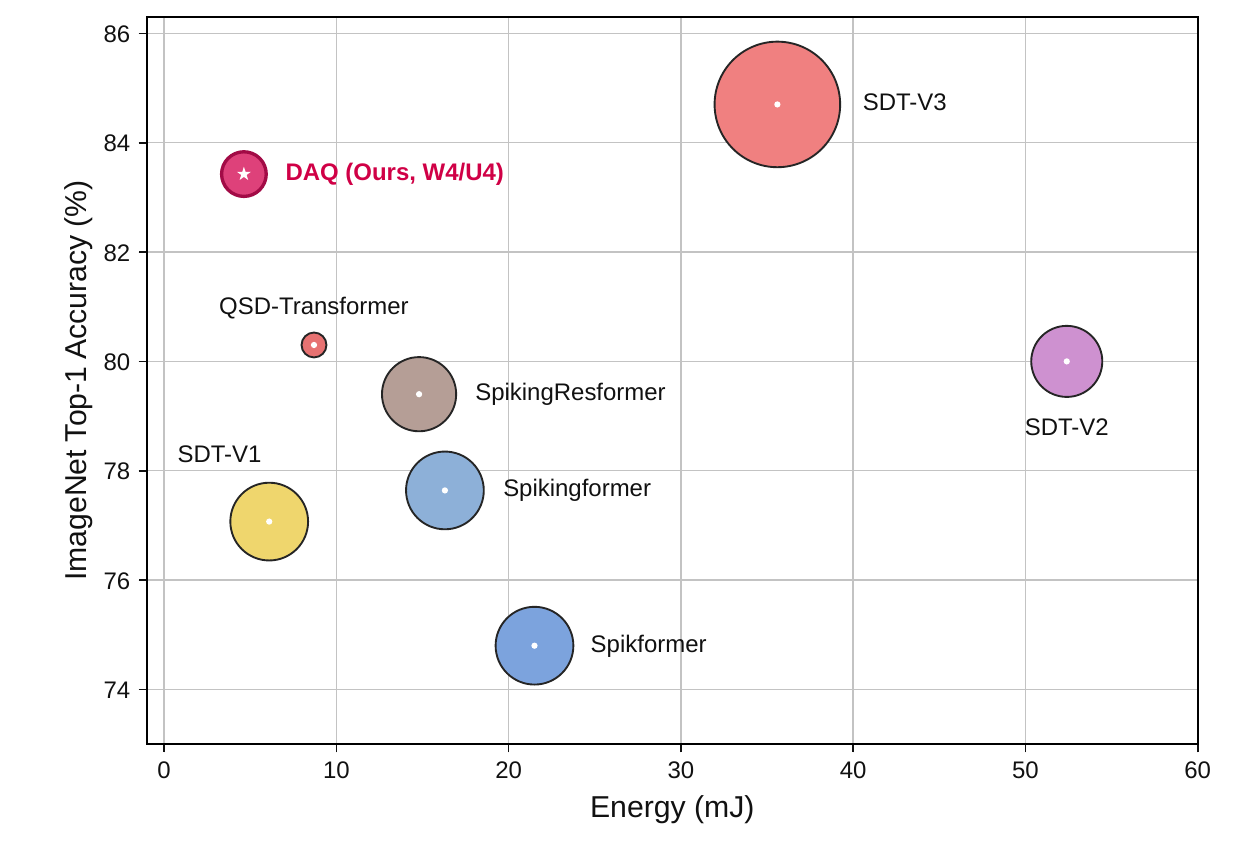}
    \caption{
        Accuracy versus energy on ImageNet-1K.
        DAQ achieves 83.4\% top-1 accuracy with 4-bit
        weights and membrane variables (W4/U4).
        Bubble area represents parameter storage.
    }
    \label{fig:imagenet-energy}
\end{figure}

\begin{figure*}[t]
    \centering
    \includegraphics[width=\linewidth]{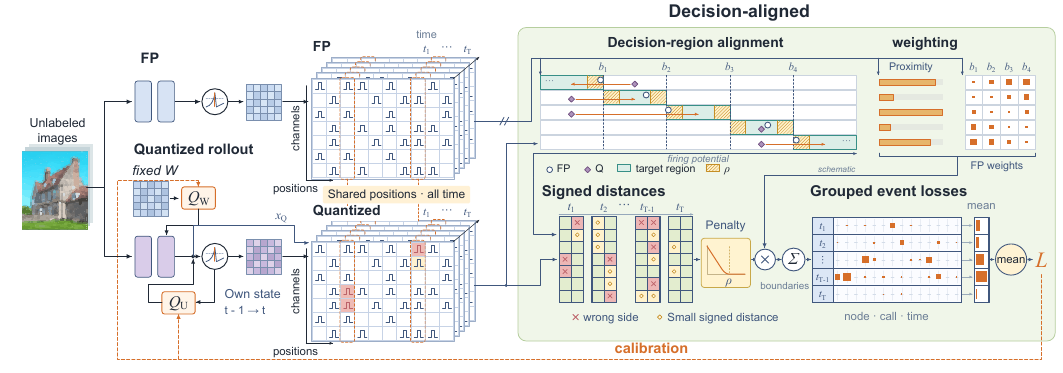}
    \caption{
    Overview of DAQ.
    The frozen floating-point model provides reference
    firing decisions, while the quantized model follows
    its own state trajectory.
    At shared sampled positions, a weighted decision-aligned
    loss encourages quantized potentials to remain within
    the reference decision regions.
    Event losses are averaged within and across groups
    to calibrate the weight and membrane quantizers.
}
    \label{fig:daq-framework}
\end{figure*}

In recurrent LIF neurons, the effect of quantization also
extends across time~\cite{ponghiran2022spiking}.
The floating-point and quantized models may both remain
silent at one step yet retain different membrane states.
These states enter the next integration step and can produce
a later firing mismatch even under identical inputs.
Weight quantization perturbs the neuronal inputs, while
membrane quantization perturbs the retained states.

To address these challenges, we propose
Decision-Aligned Post-Training Quantization (DAQ)
for spiking transformers.
DAQ calibrates weight and membrane quantizers
using the firing behavior of the floating-point model
as the target.
Guided by our analysis, we introduce a decision-aligned
objective to preserve firing decisions after quantization.
Calibration uses complete quantized forward passes,
accounting for the influence of earlier membrane states
on subsequent firing in recurrent neurons.
As illustrated in Fig.~\ref{fig:daq-framework},
DAQ provides a unified calibration framework
for neurons.

Our main contributions are summarized as follows:
\begin{itemize}
    \item
    We derive firing-preservation conditions for neurons, accounting for decision boundaries and
    the temporal propagation of state errors in recurrent LIF.

    \item
    We propose a decision-aligned objective for calibrating
    weight and membrane quantizers and derive an upper bound
    on weighted firing error in terms of the calibration loss.

    \item
    We evaluate DAQ across architectures and model scales
    on image classification, semantic segmentation, and
    event-based tracking.
    The results demonstrate competitive W4/U4 performance
    with substantial storage savings and reduced energy.
\end{itemize}

\section{Related Work}
\label{sec:related-work}

\subsection{Spiking Transformers and Neuronal Dynamics}

Spiking transformers adapt attention and residual connections
to spike-based computation.
Spikformer~\cite{zhou2022spikformer} introduces self-attention
with binary query, key, and value tensors, while
Spike-driven Transformer~\cite{yao2023spike} combines
spike-driven attention with membrane-potential residual
connections.
Subsequent architectures explore hierarchical feature
extraction through convolution--attention
backbones~\cite{yao2024spike,shi2024spikingresformer}
and efficient token interaction through
Q-K attention~\cite{zhou2024qkformer}.
These advances support event-based tracking with
SDTrack~\cite{shan2025sdtrack} and image segmentation with
Spike2Former~\cite{lei2025spike2former}.
TP-Spikformer~\cite{wei2026tp} further reduces computation
by pruning tokens according to their spatiotemporal importance.

Neuronal representations provide another dimension of
spiking transformer design.
LIF neurons integrate inputs through membrane states
retained across time, with firing and reset determining
subsequent state updates~\cite{gerstner2002spiking,ponghiran2022spiking}.
Spike firing approximation uses integer-valued activations
during training and converts them into spike trains for
inference, enabling the scaling of spike-driven
transformers~\cite{yao2025scaling}.
Membrane-potential residual connections additionally preserve
nonbinary signals within the network~\cite{yao2023spike,yao2024spike}.
Our study considers recurrent LIF neurons and a
stateless integer-valued firing configuration, retaining
the original neuronal dynamics and residual paths.

\begin{figure*}[t]
    \centering
    \includegraphics[width=\linewidth]{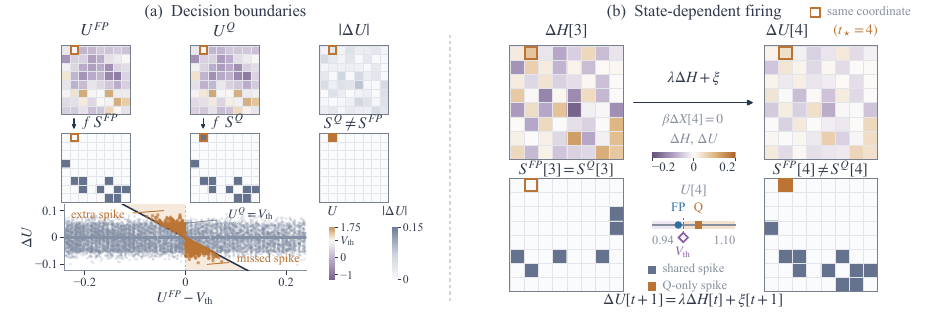}
    \caption{
        Decision boundaries and temporal propagation of
        quantization errors.
        (a) Firing preservation depends on the reference potential
        and the direction of the perturbation relative to the
        decision boundary.
        (b) Matching non-firing outputs can leave different
        post-reset states and lead to a later firing mismatch.
    }
    \label{fig:daq-analysis}
\end{figure*}

\subsection{Post-Training Quantization of Spiking Neural Networks}

Quantization-aware training (QAT) adapts SNN parameters
to low-precision representations during
training~\cite{wei2024q,qiu2025quantized}.
Q-SNNs~\cite{wei2024q} quantizes weights and membrane
potentials with weight--spike dual regulation, while
SpQuant-SNN~\cite{hasssan2024spquant} combines low-precision
membrane potentials with activation sparsity.
For spiking transformers, QSD-Transformer~\cite{qiu2025quantized}
introduces information-enhanced neurons and distillation
to train low-bit models.
HardF-SNN~\cite{liu2026hardf} further addresses hardware
execution through shared-scale quantization and integer-only
batch normalization.

Post-training quantization (PTQ) calibrates pretrained
models using limited data without full
retraining~\cite{nagel2020up,li2021brecq,yuan2022ptq4vit,ding2023towards,li2023repq,wu2024adalog,moon2024instance,yang2024dopq,jiang2026uq}.
AdaRound~\cite{nagel2020up} optimizes weight rounding through
layer reconstruction, and BRECQ~\cite{li2021brecq} uses
block reconstruction with second-order error weighting.
For vision transformers, FIMA-Q~\cite{wu2025fima} formulates
the reconstruction objective using a diagonal-plus-low-rank
approximation of the Fisher information matrix.
Precision selection has also been investigated for spiking
transformers: QSViT~\cite{putra2025qsvit} selects quantization
settings through layerwise precision analysis, while
AQ4SViT~\cite{putra2026aq4svit} uses membrane-potential drift
to screen candidates during quantization search.

Membrane-aware PTQ extends calibration to the recurrent
states that participate in subsequent neuronal
updates~\cite{asim2026flowq,xie2026ptq4snn}.
FlowQ~\cite{asim2026flowq} couples weight and membrane
scales through power-of-two ratios and calibrates them, allowing scale conversion
through bit shifts.
PTQ4SNN~\cite{xie2026ptq4snn} introduces channel-wise
scale coupling and allocates 2-, 4-, or 8-bit membrane
precision according to firing activity and quantization
sensitivity under an average-bit budget.
These approaches establish scale compatibility and precision
allocation as key components of weight and membrane
quantization~\cite{asim2026flowq,xie2026ptq4snn}.

At fixed weight and membrane bit widths, DAQ uses
neuronal decision regions to guide calibration and preserve
the firing behavior of the pretrained model.

\section{Quantization Error Analysis of Spiking Transformers}
\label{sec:daq-analysis}

Quantization can alter both the potential used for the current
firing decision and the state carried into subsequent time steps.
We analyze these effects in neurons
through their decision regions and state dynamics.

\subsection{Neuron Models and Quantization}
\label{sec:daq-setup}

Superscripts $FP$ and $Q$ denote the floating-point reference
and quantized models, respectively.
Let $U[t]$, $H[t]$, $S[t]$, and $X[t]$ denote the pre-firing
potential, post-reset state carried into the next step,
firing output, and neuron input.
For LIF with zero hard reset, the dynamics in real
arithmetic are
\begin{equation}
\begin{aligned}
U[t] &= \lambda H[t-1]+\beta X[t],\\
S[t] &= \mathbf{1}[U[t]\ge V_{\mathrm{th}}],\\
H[t] &= (1-S[t])U[t],
\end{aligned}
\label{eq:daq-neuron-dynamics}
\end{equation}
where $V_{\mathrm{th}}>0$ is the firing threshold, and
$\lambda$ and $\beta$ are the leakage and input coefficients.

For the integer-valued LIF (I-LIF) configuration,
we follow the definition in~\cite{yao2025scaling}.
We denote its firing function by $f_I$ and maximum spike
count by $D$, and analyze the integer counts before
output normalization.

The quantized model applies $Q_W$ to weights while retaining
the original charging, firing, and reset rules.
Membrane quantization $Q_U$ acts on the post-reset state
carried into the next step in LIF, and on the potential
supplied to $f_I$ in stateless I-LIF.
Throughout the analysis, $U$ denotes the potential supplied
to the firing function; for I-LIF, $U^Q$ is therefore measured
after membrane quantization.

\subsection{Decision Regions and Firing Preservation}
\label{sec:daq-decisions}

For a firing rule $f$, define the decision region
$\mathcal{D}_y=\{U:f(U)=y\}$.
At a paired neuron evaluation, let
$\Delta U=U^Q-U^{FP}$.
The firing outputs agree if and only if the quantized potential
remains in the reference decision region:
\begin{equation}
S^Q=S^{FP}
\quad\Longleftrightarrow\quad
U^{FP}+\Delta U\in\mathcal{D}_{S^{FP}}.
\label{eq:daq-region}
\end{equation}

The finite boundaries of $\mathcal{D}_{S^{FP}}$ determine the admissible range of $\Delta U$. An upper boundary limits positive perturbations, while a lower boundary limits negative perturbations.

Figure~\ref{fig:daq-analysis}(a) illustrates this relation. The line $U^{FP}+\Delta U=V_{\mathrm{th}}$ separates the quantized firing outcomes, and the reference output determines which side preserves firing. Perturbations of equal magnitude but opposite directions can therefore produce different firing outcomes. One may move the potential outside the reference decision region, while the other leaves it inside. Firing preservation thus depends on the reference potential's distance to the decision boundaries and on the direction and magnitude of the perturbation.

\subsection{Temporal Error Propagation}
\label{sec:daq-temporal}

For recurrent LIF, the two models may remain silent at the
current step yet retain different post-reset states.
These state differences can affect later firing decisions,
as illustrated in Fig.~\ref{fig:daq-analysis}(b).
Define $\Delta H[t]=H^Q[t]-H^{FP}[t]$ and
$\Delta X[t]=X^Q[t]-X^{FP}[t]$.
The next pre-firing potential difference satisfies
\begin{equation}
\Delta U[t+1]=\lambda\Delta H[t]+\xi[t+1],
\label{eq:daq-potential-error}
\end{equation}
where $\xi[t]$ contains $\beta\Delta X[t]$ and the difference
between the rounding errors of the two charging operations.
The input difference $\Delta X[t]$ includes changes caused
by both quantized weights and upstream activations.

Let $\epsilon[t]$ denote the difference between the two
models' residuals from reset arithmetic and state storage.
When the current firing outputs agree,
\begin{equation}
\Delta H[t]
=(1-S^{FP}[t])\Delta U[t]+\epsilon[t].
\label{eq:daq-matched-state}
\end{equation}
Using the input and state-storage perturbations along the
paired trajectories, define
\begin{equation}
E[1]=\lambda\Delta H[0]+\xi[1],
\label{eq:daq-conditional-initial}
\end{equation}
and, for $t\ge2$,
\begin{equation}
\begin{split}
E[t] &= \lambda(1-S^{FP}[t-1])E[t-1]\\
     &\quad+\lambda\epsilon[t-1]+\xi[t].
\end{split}
\label{eq:daq-conditional-recurrence}
\end{equation}
As long as the firing histories agree through step $t-1$,
this recurrence yields $E[t]=\Delta U[t]$.
The exact residual definitions and derivation are provided
in Appendix A.
For stateless I-LIF, we use $E[t]=\Delta U[t]$ at each
evaluation, including the perturbation introduced by
pre-firing membrane quantization.

\paragraph{Proposition 1 (Firing preservation).}
For either neuron configuration, the reference and quantized
firing outputs agree over $T$ steps if and only if
\begin{equation}
U^{FP}[t]+E[t]\in\mathcal{D}_{S^{FP}[t]},
\qquad \forall t=1,\ldots,T.
\label{eq:daq-trajectory-condition}
\end{equation}

\paragraph{Proof.}
We first show by induction that
$E[t]=\Delta U[t]$ whenever the firing histories agree
through step $t-1$.
At $t=1$, Equations~\eqref{eq:daq-potential-error}
and~\eqref{eq:daq-conditional-initial} give
$E[1]=\Delta U[1]$.
For $t\ge2$, suppose the outputs agree through step $t-1$
and $E[t-1]=\Delta U[t-1]$.
Equations~\eqref{eq:daq-potential-error}
and~\eqref{eq:daq-matched-state} then yield
\begin{equation}
\begin{aligned}
\Delta U[t]
&=\lambda\Delta H[t-1]+\xi[t]\\
&=\lambda(1-S^{FP}[t-1])E[t-1]\\
&\quad+\lambda\epsilon[t-1]+\xi[t]\\
&=E[t],
\end{aligned}
\end{equation}
where the last equality follows from
Equation~\eqref{eq:daq-conditional-recurrence}.
This establishes the conditional equality.

If the outputs agree over all $T$ steps, then
$U^{FP}[t]+E[t]=U^Q[t]$ at every step.
Equation~\eqref{eq:daq-region} therefore gives
Equation~\eqref{eq:daq-trajectory-condition}.

Conversely, assume
Equation~\eqref{eq:daq-trajectory-condition}
holds at every step.
At $t=1$, $E[1]=\Delta U[1]$, so
Equation~\eqref{eq:daq-region} gives
$S^Q[1]=S^{FP}[1]$.
If the outputs agree through step $t-1$,
the preceding result gives $E[t]=\Delta U[t]$,
and the region condition ensures
$S^Q[t]=S^{FP}[t]$.
Induction proves agreement over all $T$ steps.
\hfill$\square$

Under matched firing histories, leakage and reset determine
how state errors propagate.
During shared non-firing steps, the potential error
contributes to the next charging step with coefficient
$\lambda$.
A shared firing event clears the local saved-state difference
when $Q_U(0)=0$.
The signed input and state-storage perturbations contribute
to the same firing potential.

These interactions motivate calibration of weight and
membrane quantizers using the actual firing potentials along
the quantized model's trajectory.

\section{Method}
\label{sec:daq-method}

Figure~\ref{fig:daq-framework} summarizes DAQ.
After initializing the weight and membrane quantizers,
we optimize their parameters using the
decision-aligned objective on unlabeled calibration data.

\subsection{Quantized Representations}
\label{sec:daq-representations}

We use per-channel affine quantization for weights and
membrane variables.
For bit width $b$, scale $s>0$, and integer zero point $z$,
the quantizer is defined by
\begin{equation}
c(r)=
\operatorname{clip}\!\left(
\operatorname{round}(r/s+z),0,2^b-1
\right),
\label{eq:daq-code}
\end{equation}
with the decoded value
\begin{equation}
Q(r;s,z)=s\bigl(c(r)-z\bigr).
\label{eq:daq-affine}
\end{equation}

For LIF, $Q_U$ quantizes the post-reset state carried into
the next time step.
We learn its scale and zero point, with the grid constrained
to represent reset zero exactly and keep all decoded states
strictly below $V_{\mathrm{th}}$ in the source precision.

For I-LIF, $Q_U$ quantizes the pre-firing potential. We calibrate the weight and membrane quantizers using a firing-alignment objective defined against the floating-point reference.

\subsection{Decision-Aligned Calibration}
\label{sec:daq-objective}

For each paired neuron evaluation $e$, let $y_e$ and $q_e$
denote the reference and quantized firing outputs before
normalization.
We use $x_e=U_e/V_{\mathrm{th}}$ for LIF and $x_e=U_e$
for I-LIF.
The quantized I-LIF potential is read after $Q_U$ and
before applying $f_I$.
Let $\mathcal{J}_e$ denote the index set of firing decision
boundaries for neuron evaluation $e$, with $b_j$ denoting
the location of boundary $j$.

The reference output specifies the required side of each
boundary.
We define the signed distance
\begin{equation}
d_{ej}=
\begin{cases}
x_e^Q-b_j, & y_e\ge j,\\
b_j-x_e^Q, & y_e<j.
\end{cases}
\label{eq:daq-signed-distance}
\end{equation}
Positive values place the quantized potential on the
reference side, while negative values place it on the
opposite side.
We penalize distances below $\rho>0$ to encourage
separation from the boundary.

An I-LIF firing error can span several spike counts.
To reflect its magnitude, we assign each boundary the weight
\begin{equation}
\omega_j(y_e)=|2j-1-2y_e|.
\label{eq:daq-boundary-weight}
\end{equation}
These weights sum to the squared count error over
disagreeing boundaries, as shown below.

A small perturbation can change firing when the reference
potential lies near a boundary.
We emphasize these events through the weight
$1/(\mu_e+\delta)$, where $\mu_e$ is the reference distance
and $\delta>0$ keeps the weight finite.
For LIF, $\mu_e=|x_e^{FP}-1|$.
For I-LIF, $\mu_e=
\min_{j\in\mathcal{J}_e}
\left|
\operatorname{clip}(x_e^{FP},0,D)-b_j
\right|$.
The resulting event loss is
\begin{equation}
\ell_e=
\frac{1}{\mu_e+\delta}
\sum_{j\in\mathcal{J}_e}
\omega_j(y_e)
H_+\!\left(\frac{\rho-d_{ej}}{\eta}\right),
\label{eq:daq-event-loss}
\end{equation}
where $H_+(r)$ is a one-sided Huber penalty:
it is zero for $r\le0$, equals $r^2/2$ for $0<r<1$,
and equals $r-\tfrac12$ for $r\ge1$.
The quadratic branch provides a smooth transition to zero,
while the linear branch maintains a constant slope for
larger violations.
The parameter $\eta$ controls the transition width.
We use $\rho=\eta=0.25$ and $\delta=0.05$, treating
$y_e$ and $\mu_e$ as constants when computing gradients.
The loss is zero exactly when $d_{ej}\ge\rho$ for every
boundary.

\paragraph{Proposition 2 (Firing-error bound).}
For either neuron configuration and $\rho=\eta>0$,
each paired evaluation satisfies
\begin{equation}
\frac{(q_e-y_e)^2}{\mu_e+\delta}\le 2\ell_e.
\label{eq:daq-event-bound}
\end{equation}
For LIF, the squared difference is the spike-mismatch
indicator; for I-LIF, it is the squared count error.

\paragraph{Proof.}
Let $\chi_{ej}$ indicate whether exactly one of $q_e$
and $y_e$ is at least $j$.
The boundary weights satisfy
\begin{equation}
(q_e-y_e)^2=
\sum_{j\in\mathcal{J}_e}
\omega_j(y_e)\chi_{ej}.
\label{eq:daq-ordinal-identity}
\end{equation}
Only boundaries between the two outputs contribute.
Their weights are $1,3,\ldots,2|q_e-y_e|-1$,
which sum to $(q_e-y_e)^2$.
At each disagreeing boundary, $d_{ej}\le0$.
Since $\rho=\eta$, the corresponding Huber term is
at least $H_+(1)=1/2$.
Multiplying by $\omega_j(y_e)$, summing over boundaries,
and dividing by $\mu_e+\delta$ proves
Equation~\eqref{eq:daq-event-bound}.
\hfill$\square$

We group sampled events by node, invocation, and time step,
with $\mathcal{E}_g$ denoting the events in group $g$.
Averaging within each group and equally across the $G$
groups gives
\begin{equation}
L=
\frac{1}{G}\sum_{g=1}^{G}
\frac{1}{|\mathcal{E}_g|}
\sum_{e\in\mathcal{E}_g}\ell_e.
\label{eq:daq-aggregation}
\end{equation}
Let $R_w$ denote the same group average of
$(q_e-y_e)^2/(\mu_e+\delta)$.
Proposition~2 then gives $R_w\le2L$.
Algorithm~\ref{alg:daq-calibration} summarizes the
calibration procedure.

\newcounter{daqalgorithm}
\begin{figure}[t]
\centering
\begin{minipage}{\columnwidth}
\small
\refstepcounter{daqalgorithm}
\label{alg:daq-calibration}
\hrule
\smallskip
\noindent
\textbf{Algorithm \thedaqalgorithm}
Decision-Aligned Quantization.
\par\smallskip
\hrule
\smallskip
\begin{enumerate}[
    label=\scriptsize\arabic*:,
    leftmargin=1.8em,
    labelsep=0.45em,
    align=right,
    nosep
]
    \item[] \textbf{Input:}
    Reference model $\mathcal{M}^{FP}$,
    unlabeled calibration set $\mathcal{X}$.
    \item[] \textbf{Output:}
    Quantized model $\mathcal{M}^{Q}$.

    \item[] \textcolor[RGB]{49,145,137}{\# Initialization}
    \item Initialize $Q_W$ and $Q_U$ under the grid constraints.

    \item[] \textcolor[RGB]{49,145,137}{\# Decision alignment}
    \item \textbf{for} each calibration update \textbf{do}

    \item \hspace*{1em}%
    \parbox[t]{\dimexpr\linewidth-1em\relax}{%
        Select a batch and uniformly sample positions
        shared across channels and time.}

    \item \hspace*{1em}%
    \parbox[t]{\dimexpr\linewidth-1em\relax}{%
        Obtain reference targets $(y_e,\mu_e)$
        at the sampled positions.}

    \item \hspace*{1em}%
    \parbox[t]{\dimexpr\linewidth-1em\relax}{%
        Run the full quantized forward pass using its
        own states and compute $L$ over the sampled
        events with Equation~\eqref{eq:daq-aggregation}.}

    \item \hspace*{1em}%
    \parbox[t]{\dimexpr\linewidth-1em\relax}{%
        Update the parameters of $Q_W$ and $Q_U$
        using local surrogate gradients, retaining
        temporal state gradients within nodes and
        detaching firing outputs between nodes.}

    \item \hspace*{1em}%
    \parbox[t]{\dimexpr\linewidth-1em\relax}{%
        Project the quantizers onto their grid constraints.}

    \item \textbf{end for}
    \item Fix the calibrated quantizer parameters
    and return $\mathcal{M}^{Q}$.
\end{enumerate}
\smallskip
\hrule
\end{minipage}
\end{figure}

\begin{table*}[t]
\centering
\caption{
Event-based object tracking results on FE108, FELT, and VisEvent.
Ratios beside TP-Spikformer indicate average token retention.
The three best distinct values per metric are highlighted as
\daqfirst{first}, \daqsecond{second}, and \daqthird{third}.
Ties share the same highlight; a dash denotes an unavailable result. $T$ denotes the number of time steps, and $D$ denotes the
maximum spike count per time step.
}
\label{tab:event_tracking}

\small
\setlength{\tabcolsep}{2pt}
\renewcommand{\arraystretch}{1.08}

\begin{tabular*}{\textwidth}{
    @{\extracolsep{\fill}} l c c r *{6}{r} @{}
}
\toprule
Method
& $T\times D$
& Precision
& \multicolumn{1}{c}{Energy}
& \multicolumn{2}{c}{FE108}
& \multicolumn{2}{c}{FELT}
& \multicolumn{2}{c}{VisEvent} \\
\cmidrule(lr){5-6}
\cmidrule(lr){7-8}
\cmidrule(lr){9-10}
& &
& \multicolumn{1}{c}{(mJ) $\downarrow$}
& \multicolumn{1}{c}{AUC (\%) $\uparrow$}
& \multicolumn{1}{c}{PR (\%) $\uparrow$}
& \multicolumn{1}{c}{AUC (\%) $\uparrow$}
& \multicolumn{1}{c}{PR (\%) $\uparrow$}
& \multicolumn{1}{c}{AUC (\%) $\uparrow$}
& \multicolumn{1}{c}{PR (\%) $\uparrow$} \\
\midrule

STARK~\cite{yan2021learning}
& $1{\times}1$ & FP & 58.88
& 57.4 & 89.2
& \daqfirst{39.6} & \daqfirst{51.7}
& 34.1 & 46.8 \\

SimTrack~\cite{chen2022backbone}
& $1{\times}1$ & FP & 93.84
& 56.7 & 88.3
& 36.8 & 47.0
& 34.6 & 47.6 \\

OSTrack$_{256}$~\cite{ye2022joint}
& $1{\times}1$ & FP & 98.90
& 54.6 & 87.1
& 35.9 & 45.5
& 32.7 & 46.4 \\

ARTrack$_{256}$~\cite{wei2023autoregressive}
& $1{\times}1$ & FP & 174.8
& 56.6 & 88.5
& \daqsecond{39.5} & 49.4
& 33.0 & 43.8 \\

SeqTrack-B$_{256}$~\cite{chen2023seqtrack}
& $1{\times}1$ & FP & 302.7
& 53.5 & 85.5
& 33.0 & 42.0
& 28.6 & 43.3 \\

HIPTrack~\cite{cai2024hiptrack}
& $1{\times}1$ & FP & 307.7
& 50.8 & 81.0
& 38.2 & 48.9
& 32.1 & 45.2 \\

ODTrack~\cite{zheng2024odtrack}
& $1{\times}1$ & FP & 335.8
& 43.2 & 69.7
& 29.7 & 35.9
& 24.7 & 34.7 \\

\addlinespace[3pt]
STNet~\cite{zhang2022spiking}
& $3{\times}1$ & FP & 103.53
& -- & --
& -- & --
& 35.0 & \daqsecond{50.3} \\

SNNTrack~\cite{zhang2025spiking}
& $5{\times}1$ & FP & 8.25
& -- & --
& -- & --
& 35.2 & \daqfirst{50.4} \\

\midrule
SDTrack$_{\mathrm{Tiny}}$~\cite{shan2025sdtrack}
& $4{\times}1$ & FP & 8.15
& 56.7 & 89.1
& 35.8 & 44.1
& \daqsecond{35.4} & 48.7 \\

& $1{\times}4$ & FP & 8.16
& \daqfirst{59.0} & \daqfirst{91.3}
& \daqsecond{39.3} & \daqsecond{51.2}
& \daqfirst{35.6} & 49.2 \\

\midrule
\addlinespace[3pt]
TP-Spikformer ($\times0.65$)~\cite{wei2026tp}
& $1{\times}4$ & FP & 6.65
& \daqfirst{59.0} & \daqsecond{91.2}
& \daqthird{39.1} & \daqthird{50.4}
& \daqthird{35.3} & \daqthird{49.7} \\

TP-Spikformer ($\times0.56$)
& $1{\times}4$ & FP & \daqthird{6.51}
& \daqthird{58.4} & 90.6
& 38.9 & 50.0
& 35.2 & 49.4 \\

\midrule
\textbf{DAQ (Ours)}
& $4{\times}1$ & W4/U4 & \daqsecond{2.91}
& 56.6 & 88.7
& 34.8 & 42.8
& 35.0 & 48.2 \\

& $1{\times}4$ & W4/U4 & \daqfirst{2.76}
& \daqsecond{58.8} & \daqthird{90.9}
& 38.1 & 48.8
& 34.5 & 48.0 \\

\bottomrule
\end{tabular*}
\end{table*}

\section{Experiments}
\label{sec:experiments}

We evaluate DAQ under W4/U4 quantization on event-based
object tracking, image classification, and semantic
segmentation across different architectures and model scales.
We compare task performance and resource costs with
floating-point counterparts and existing methods.
We then examine the calibration design through ablation
studies and report the wall-clock quantization time.

\paragraph{Experimental Setup.}
We calibrate weight and membrane quantizers using unlabeled
data while keeping the pretrained weights fixed.
Both use per-channel affine quantization, as described in
Sec.~\ref{sec:daq-representations}.
Throughout the experiments, W$b_W$/U$b_U$ denotes
$b_W$-bit weights and $b_U$-bit membrane variables.
The datasets, model configurations, and evaluation protocols
are specified in the corresponding subsections.

\subsection{Event-Based Object Tracking}
\label{sec:daq-tracking}

We evaluate DAQ on FE108, FELT, and VisEvent using
SDTrack$_{\mathrm{Tiny}}$~\cite{shan2025sdtrack}.
We consider LIF neurons with $T\times D=4\times1$
and I-LIF neurons with $T\times D=1\times4$,
quantizing both weights and membrane variables to 4 bits.
Tracking performance is measured by the area under the
success curve (AUC) and precision rate (PR).

As shown in Table~\ref{tab:event_tracking}, DAQ reduces energy to 2.91~mJ for LIF and 2.76~mJ for I-LIF, corresponding to reductions of 64.3\% and 66.2\% relative to their floating-point counterparts.
Across the three benchmarks, the quantized LIF model stays within
1.0 percentage point of its floating-point reference in AUC and
within 1.3 points in PR.
For I-LIF, DAQ reaches 58.8\% AUC and 90.9\% PR on FE108,
giving up 0.2 points in AUC and 0.4 points in PR.
AUC falls by 1.2 points on FELT and by 1.1 points on VisEvent.

DAQ also consumes the least energy among the compared methods.
Relative to TP-Spikformer ($\times0.56$)~\cite{wei2026tp},
the quantized I-LIF model cuts energy use by 57.6\%
while gaining 0.4 points in AUC and 0.3 points in PR on FE108.
It gives up 0.8 points of AUC on FELT and 0.7 points on VisEvent.

\begin{table*}[t]
\centering
\caption{
ImageNet-1K classification results.
Parameter storage is expressed in millions of FP32-equivalent
parameters. $^\ast$ denotes results reproduced by us
with W4/U4 adaptations of the official implementations. 
}
\label{tab:main_ptq_results}

\small
\setlength{\tabcolsep}{2pt}
\renewcommand{\arraystretch}{1.08}

\begin{tabular*}{\textwidth}{
    @{\extracolsep{\fill}}
    l c c r r r
    @{}
}
\toprule
Method
& Bits (W/U)
&  $T\times D$
& \multicolumn{1}{c}{Param.\ (M) }
& \multicolumn{1}{c}{Energy (mJ) }
& \multicolumn{1}{c}{Top-1 (\%)} \\
\midrule

Spikformer~\cite{zhou2022spikformer}
& 32/32 & $4{\times}1$
& 66.3 & 21.5 & 74.8 \\

SDT-V1~\cite{yao2023spike}
& 32/32 & $4{\times}1$
& 66.3 & 6.1 & 77.1 \\

Spikingformer~\cite{zhou2026spikingformer}
& 32/32 & $4{\times}1$
& 66.3 & 16.3 & 77.6 \\

SpikingResformer~\cite{shi2024spikingresformer}
& 32/32 & $4{\times}1$
& 60.4 & 14.8 & 79.4 \\

SNN-ViT~\cite{wang2025spiking}
& 32/32 & $4{\times}1$
& 53.7 & 35.8 & 80.2 \\

\midrule

SDT-V2~\cite{yao2024spike}
& 32/32 & $4{\times}1$
& 55.4 & 52.4 & 79.7 \\

\midrule

QSD-Transformer~\cite{qiu2025quantized}
& 4/32 & $4{\times}1$
& \daqsecond{6.8} & 8.7 & 80.3 \\

\midrule

SDT-V3~\cite{yao2025scaling}
& 32/32 & $1{\times}4$
& 19.0 & 5.9 & 79.8 \\

& 32/32 & $1{\times}4$
& 83.0 & 19.1 & \daqthird{83.2} \\

& 32/32 & $1{\times}4$
& 171.0 & 35.6 & \daqfirst{84.7} \\

\midrule

TP-Spikformer ($\times0.78$)~\cite{wei2026tp}
& 32/32 & $1{\times}4$
& 19.0 & 4.7 & 79.0 \\

TP-Spikformer ($\times0.56$)
& 32/32 & $1{\times}4$
& 19.0 &4.3 & 77.6 \\

\midrule

BRECQ$^\ast$~\cite{li2021brecq}
& 4/4 & $1{\times}4$
& 2.5 & \daqsecond{1.3} & 76.9 \\

\midrule
DAQ (Ours)
& 4/4 & $1{\times}4$
& \daqfirst{2.5} & \daqfirst{1.2} & 77.7 \\

& 4/4 & $1{\times}4$
& \daqthird{10.7} &  \daqthird{2.5} & 82.0 \\

& 4/4 & $1{\times}4$
& 21.9 & 4.6 & \daqsecond{83.4} \\

\bottomrule
\end{tabular*}
\end{table*}

\begin{table}[t]
\centering
\caption{
Top-1 accuracy (\%) on CIFAR-10 and CIFAR-100 at $T=4$.
$^\dagger$ uses 4-bit weights and mixed-precision
membrane quantization under an average 4-bit membrane budget. Numbers in method names denote token retention ratios.
}
\label{tab:cifar_results}

\small
\setlength{\tabcolsep}{2pt}
\renewcommand{\arraystretch}{1.08}

\begin{tabular*}{\columnwidth}
{@{\extracolsep{\fill}}l c r r @{}}
\toprule
Method & Bits & CIFAR-10 & CIFAR-100 \\
\midrule

\begin{tabular}[t]{@{}l@{}}
SparseSpikformer
($\times0.63$)~\cite{liu2024sparsespikformer}
\end{tabular}
& 32/32 & 94.8 & 76.8 \\

\begin{tabular}[t]{@{}l@{}}
AT-SNN
($\times0.28$)~\cite{kang2024snn}
\end{tabular}
& 32/32 & \daqthird{95.1} & \daqsecond{78.1} \\

\begin{tabular}[t]{@{}l@{}}
STATA
($\times0.50$)~\cite{zhuge2024towards}
\end{tabular}
& 32/32 & 95.0 & 77.7 \\

\begin{tabular}[t]{@{}l@{}}
TP-Spikformer
($\times0.25$)~\cite{wei2026tp}
\end{tabular}
& 32/32 & \daqsecond{95.2} & \daqfirst{78.5} \\

\midrule
BRECQ$^\ast$~\cite{li2021brecq}
& 4/4 & 94.1 & 73.2 \\

PTQ4SNN$^{\dagger}$~\cite{xie2026ptq4snn}
& 4/4 & 94.4 & 75.3 \\

\midrule
DAQ (Ours)
& 4/4 & \daqfirst{95.3} & \daqthird{77.8} \\

\bottomrule
\end{tabular*}
\end{table}
\subsection{Image Classification}
\label{sec:daq-classification}

We evaluate DAQ on ImageNet-1K, CIFAR-10, and CIFAR-100
using top-1 accuracy.
All DAQ models use W4/U4 quantization.

\paragraph{ImageNet-1K.}
We apply DAQ to pretrained SDT-V3-19M, SDT-V3-83M,
and SDT-V3-171M models~\cite{yao2025scaling}
with $T\times D=1\times4$.
We also adapt BRECQ~\cite{li2021brecq} to SDT-V3-19M
under W4/U4 using their official implementations.
For BRECQ, we use block reconstruction to optimize
weight rounding and membrane scales.

Table~\ref{tab:main_ptq_results} shows that DAQ preserves
the accuracy advantage of larger backbones under
low-bit quantization.
The 83M and 171M models remain within 1.3 percentage
points of their floating-point counterparts,
while the 19M model incurs a decrease of 2.1 points
and outperforms our reproduced BRECQ baseline by 0.8 points.
Across these models, energy is reduced by 79.7\%--87.1\%.

These energy savings also make a larger quantized
backbone competitive with a smaller floating-point model.
The quantized SDT-V3-171M achieves 83.4\% accuracy
at 4.6~mJ, slightly exceeding the 83.2\% accuracy
of floating-point SDT-V3-83M at 19.1~mJ.
At a fixed backbone size, DAQ also provides a favorable
accuracy--energy tradeoff relative to token pruning.
On SDT-V3-19M, DAQ achieves 77.7\% accuracy,
compared with 77.6\% for TP-Spikformer
($\times0.56$)~\cite{wei2026tp},
while reducing energy from 4.3 to 1.2~mJ.

\paragraph{CIFAR-10 and CIFAR-100.}
To evaluate DAQ with recurrent membrane dynamics,
we quantize pretrained Spike-driven Transformers
(SDT-2-512)~\cite{yao2023spike} with LIF neurons
and $T=4$ on both datasets.

As shown in Table~\ref{tab:cifar_results},
DAQ achieves 95.3\% and 77.8\% accuracy on
CIFAR-10 and CIFAR-100, respectively.
It also remains competitive with the listed
floating-point token-pruning methods.
On CIFAR-10, DAQ achieves the highest accuracy
among the compared methods.
On CIFAR-100, it is 0.7 percentage points below
TP-Spikformer while using 4-bit weights
and recurrent membrane states.

\subsection{Semantic Segmentation}
\label{sec:daq-segmentation}

\begin{table}[t]
\centering
\caption{
Semantic segmentation results on ADE20K. 
}
\label{tab:ade20k_segmentation}
\footnotesize
\setlength{\tabcolsep}{1.5pt}
\renewcommand{\arraystretch}{1.12}

\begin{tabular*}{\columnwidth}
{@{\extracolsep{\fill}}lccrrr@{}}
\toprule
Method & $T{\times}D$ & Bits & Param. & Energy & mIoU \\
\midrule
SDT-V2~\cite{yao2024spike}
& $4{\times}1$ & FP & 16.5 & 88.1 & 33.6 \\
& $4{\times}1$ & FP & 59.8 & 183.6 & 35.3 \\

\addlinespace[3pt]
SDT-V3~\cite{yao2025scaling}
& $1{\times}4$ & FP & 11.0
& \daqthird{27.2} & \daqthird{40.1} \\

\addlinespace[3pt]
QSD (V2-T)~\cite{qiu2025quantized}
& $4{\times}1$ & W4
& \daqsecond{3.3} & \daqsecond{17.5} & 39.0 \\
QSD (V2-M)
& $4{\times}1$ & W4
& \daqthird{9.6} & 37.9 & \daqfirst{40.5} \\

\addlinespace[3pt]
TP-Spikformer ($\times0.78$)~\cite{wei2026tp}
& $1{\times}4$ & FP & 11.0 & -- & \daqsecond{40.2} \\
TP-Spikformer ($\times0.56$)
& $1{\times}4$ & FP & 11.0 & -- & 40.0 \\

\midrule
\textbf{DAQ (Ours)}
& $1{\times}4$ & W4/U4
& \daqfirst{3.0} & \daqfirst{15.2} & 38.9 \\
\bottomrule
\end{tabular*}

\end{table}

We evaluate DAQ on ADE20K using the pretrained
SDT-V3-10M backbone~\cite{yao2025scaling} with an FPN neck
and a segmentation head.
We apply W4/U4 quantization to the backbone weights and
membrane variables and measure segmentation performance
by mIoU.

As shown in Table~\ref{tab:ade20k_segmentation},
backbone quantization provides substantial parameter
storage savings in the segmentation model.
DAQ achieves 38.9\% mIoU, a decrease of 1.2 percentage
points from the floating-point baseline, while reducing
parameter storage from 11.0M to 3.0M FP32 equivalents.
The energy is also 44.1\% lower,
at 15.2~mJ per image compared with 27.2~mJ
for the floating-point baseline.

\begin{figure}[t]
    \centering
    \includegraphics[width=\columnwidth]{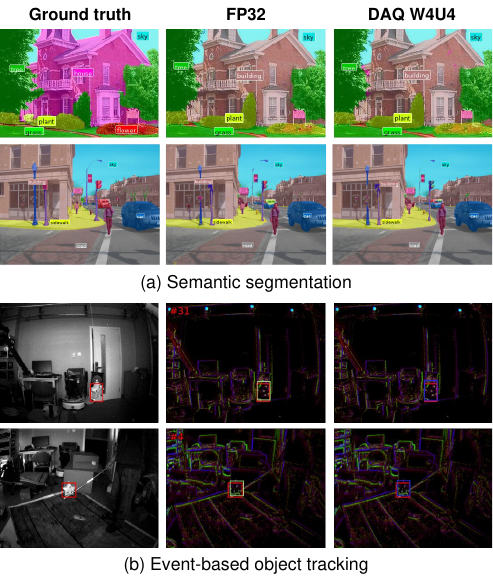}
    \caption{
        Qualitative comparison of FP32 and DAQ (W4/U4)
        for (a) semantic segmentation and (b) event-based
        object tracking. Ground truth is shown for reference.
    }
    \label{fig:daq-qualitative}
\end{figure}

Figure~\ref{fig:daq-qualitative} presents qualitative
comparisons between FP32 and DAQ for semantic
segmentation and event-based object tracking.

\subsection{Ablation Studies}
\label{sec:daq-ablation}

\begin{table}[t]
\centering
\caption{
Comparison of calibration objectives on SDT-V3-10M.
$L$ is the DAQ loss, $R_w$ is the weighted firing error,
and $R$ is the unweighted squared count error.
}
\label{tab:daq-firing-error-comparison}
\small
\setlength{\tabcolsep}{3pt}
\begin{tabular*}{\columnwidth}{
    @{\extracolsep{\fill}} l r r r r @{}
}
\toprule
Objective
& $L$ $\downarrow$
& $R_w$ $\downarrow$
& $R$ $\downarrow$
& Top-1 (\%)  \\
\midrule
MSE
& 5.83
& 1.79
& 0.39
& $74.05 \pm 0.16$ \\
DAQ
& \textbf{5.46}
& \textbf{1.70}
& \textbf{0.36}
& $\mathbf{75.98} \pm 0.11$ \\
\bottomrule
\end{tabular*}
\end{table}

We compare membrane reconstruction MSE with DAQ's
decision-aligned objective while keeping all other
calibration settings fixed.
As shown in Table~\ref{tab:daq-firing-error-comparison},
replacing MSE with the decision-aligned objective improves
mean top-1 accuracy from 74.05\% to 75.98\%
and reduces both weighted and unweighted squared count errors.
The accuracy improvement is thus accompanied by
closer agreement with the reference firing outputs.

Table~\ref{tab:calibration_objective_ablation} further
separates the contributions of the calibration components,
with Initial grids reporting accuracy before quantizer optimization.
DAQ without distance weighting sets the reference-distance
weights to one, leaving the rest of the objective unchanged.
On SDT-V3-10M, the accuracy gain over MSE largely
persists in this setting; distance weighting adds
0.08 percentage points to the mean accuracy.
Its contribution is larger on recurrent SDT,
where the gain is 0.65 points.
DAQ with reference states uses quantized reference states
during calibration and its own recurrent states during evaluation.
Calibrating along the quantized model's own state trajectory
improves mean SDT accuracy by 0.23 points over this variant,
supporting calibration with the state evolution used at inference.

\begin{table}[t]
\centering
\caption{
Ablation study on ImageNet-1K under W4/U4 quantization.
Values are reported as mean $\pm$ sample standard deviation.
}
\label{tab:calibration_objective_ablation}
\small
\setlength{\tabcolsep}{3pt}
\begin{tabular}{@{}llr@{}}
\toprule
Model & Configuration & Top-1 (\%)  \\
\midrule
SDT-8-768
& Initial grids
& $61.87\pm0.27$ \\
& DAQ without distance weighting
& $73.32\pm0.24$ \\
& DAQ with reference states
& $73.74\pm0.04$ \\
& Full DAQ
& $\mathbf{73.97}\pm0.10$ \\
\midrule
SDT-V3-10M
& Initial grids
& $68.44\pm0.17$ \\
& DAQ without distance weighting
& $75.90\pm0.08$ \\
& Full DAQ
& $\mathbf{75.98}\pm0.11$ \\
\bottomrule
\end{tabular}
\end{table}

\begin{table}[t]
\centering
\caption{
Packed parameter and membrane buffer storage at
batch size one, and W4/U4 quantization time on ImageNet-1K.
$W_{\mathrm{FP}}$ and $U_{\mathrm{FP}}$ denote the original
floating-point precisions.
}
\label{tab:daq_storage_time}
\small
\setlength{\tabcolsep}{3pt}
\begin{tabular*}{\columnwidth}{@{\extracolsep{\fill}}lrrrr@{}}
\toprule
& \multicolumn{3}{c}{Storage (MiB)} & DAQ time \\
\cmidrule(lr){2-4}
Backbone
& $W_{\mathrm{FP}}/U_{\mathrm{FP}}$
& W4/$U_{\mathrm{FP}}$
& W4/U4
& (GPU min) \\
\midrule
SDT-V3-10M  &  40.5 &  7.3 &  5.6 &  5.9 \\
SDT-V3-19M  &  75.5 & 12.4 & 10.1 &  6.3 \\
SDT-V3-171M & 664.5 & 92.7 & 85.7 &  8.3 \\
\midrule
SDT-8-768   & 314.3 & 93.4 & 47.5 & 19.6 \\
\bottomrule
\end{tabular*}
\end{table}

\subsection{Storage and Quantization Time}
\label{sec:daq-storage-time}

Table~\ref{tab:daq_storage_time} shows that membrane
quantization provides a substantial storage benefit
beyond weight quantization for recurrent SDT-8-768.
At fixed W4, U4 reduces the combined parameter and
membrane buffer storage from 93.4 to 47.5~MiB,
nearly halving the footprint.
This reflects the need to retain a separate recurrent
state for each neuron node.

For SDT-V3, the storage estimate assumes
reuse of a single membrane buffer across layers.
Its contribution to total storage becomes smaller
as parameter storage grows, so the relative benefit
of U4 decreases with model size.
The complete quantization procedure takes
5.9--8.3 minutes for SDT-V3 and 19.6 minutes
for SDT-8-768 on a single NVIDIA RTX 4090.

\section{Conclusion}

We presented DAQ, a post-training quantization method
for spiking transformers. Guided by the analysis of firing boundaries and
state propagation, DAQ uses firing preservation as a
shared calibration objective along quantized trajectories.
Experiments on image classification, semantic segmentation,
and event-based tracking demonstrate competitive performance
under W4/U4 quantization, with substantial storage savings
and reduced energy.

\section*{Acknowledgments}
This work was supported in part by the National Natural Science Foundation of China under grant 62304203 and in partby the International Campus of ZJU International Research Collaboration Seed Project under grant GJ202509.
{
    \small
    \bibliographystyle{ieeenat_fullname}
    \bibliography{main}
}
\clearpage

\maketitlesupplementary
\appendix

\section{Additional Analysis and Proofs}
\label{sec:daq-analysis-appendix}

We provide the exact decision regions, derive the error
recurrences, and prove Propositions~1 and~2.
Notation follows the main paper, with neuron and channel
indices omitted.

\subsection{Exact Decision Regions}
\label{sec:daq-app-decision-regions}

For LIF, the firing rule
$S=\mathbf{1}[U\ge V_{\mathrm{th}}]$ gives
\begin{equation}
\begin{aligned}
\mathcal{D}_0&=(-\infty,V_{\mathrm{th}}),\\
\mathcal{D}_1&=[V_{\mathrm{th}},\infty).
\end{aligned}
\label{eq:daq-app-lif-regions}
\end{equation}
Thus, firing preservation requires
$\Delta U<V_{\mathrm{th}}-U^{FP}$ when $S^{FP}=0$,
and $\Delta U\ge V_{\mathrm{th}}-U^{FP}$ when $S^{FP}=1$.
Equality belongs to the firing region.

The admissible perturbations are obtained by translating
$\mathcal{D}_{S^{FP}}$ by $-U^{FP}$:
\begin{equation}
S^Q=S^{FP}
\quad\Longleftrightarrow\quad
U^{FP}+\Delta U\in\mathcal{D}_{S^{FP}}.
\label{eq:daq-app-local-preservation}
\end{equation}
\subsection{Error Recurrences and Firing Preservation}
\label{sec:daq-app-proposition-one}

\paragraph{Arithmetic and storage residuals.}
For $m\in\{FP,Q\}$, define residuals from the realized
charging and state-update operations:
\begin{equation}
\begin{aligned}
r_U^m[t]
&=U^m[t]-\lambda H^m[t-1]-\beta X^m[t],\\
r_H^m[t]
&=H^m[t]-(1-S^m[t])U^m[t].
\end{aligned}
\label{eq:daq-app-arithmetic-residuals}
\end{equation}
The charging residual captures arithmetic rounding.
The state-update residual includes reset arithmetic and
state storage, including membrane quantization in
the quantized model.
The perturbations used in the main paper are
\begin{equation}
\begin{aligned}
\xi[t]
&=\beta\Delta X[t]+r_U^Q[t]-r_U^{FP}[t],\\
\epsilon[t]
&=r_H^Q[t]-r_H^{FP}[t].
\end{aligned}
\label{eq:daq-app-residual-differences}
\end{equation}
Both are evaluated along the paired model trajectories.
In particular, $\Delta X[t]$ includes changes caused
by quantized weights and upstream activations.

In exact arithmetic apart from quantization,
$\xi[t]=\beta\Delta X[t]$, and
\begin{equation}
\begin{split}
\epsilon[t]
&=Q_U\bigl((1-S^Q[t])U^Q[t]\bigr)\\
&\quad-(1-S^Q[t])U^Q[t].
\end{split}
\label{eq:daq-app-storage-error}
\end{equation}
Equation~\eqref{eq:daq-app-residual-differences}
also accounts for the source execution precision.

\paragraph{Error recurrence.}
Subtracting the two charging equations gives
\begin{equation}
\Delta U[t]=\lambda\Delta H[t-1]+\xi[t].
\label{eq:daq-app-potential-error}
\end{equation}
Let $\Delta S[t]=S^Q[t]-S^{FP}[t]$.
Subtracting the state updates and rearranging yields
\begin{equation}
\begin{split}
\Delta H[t]
&=(1-S^Q[t])U^Q[t]\\
&\quad-(1-S^{FP}[t])U^{FP}[t]+\epsilon[t]\\
&=(1-S^{FP}[t])\Delta U[t]\\
&\quad-U^Q[t]\Delta S[t]+\epsilon[t].
\end{split}
\label{eq:daq-app-state-error}
\end{equation}
When the outputs agree, $\Delta S[t]=0$, so
\begin{equation}
\Delta H[t]
=(1-S^{FP}[t])\Delta U[t]+\epsilon[t].
\label{eq:daq-app-matched-state}
\end{equation}
During a shared silent step, the potential difference
and storage perturbation remain in the saved state.
A shared firing event clears the saved-state difference
under zero hard reset and $Q_U(0)=0$.
The next potential difference then equals $\xi[t+1]$.

Using the paired perturbations, define
\begin{equation}
E[1]=\lambda\Delta H[0]+\xi[1],
\label{eq:daq-app-conditional-initial}
\end{equation}
and, for $t\ge2$,
\begin{equation}
\begin{split}
E[t]
&=\lambda(1-S^{FP}[t-1])E[t-1]\\
&\quad+\lambda\epsilon[t-1]+\xi[t].
\end{split}
\label{eq:daq-app-conditional-recurrence}
\end{equation}
This is the conditional recurrence used in Proposition~1.

\paragraph{Proof of Proposition 1.}
For LIF, Equation~\eqref{eq:daq-app-potential-error}
gives $E[1]=\Delta U[1]$.
Suppose the firing histories agree through step $t-1$
and $E[t-1]=\Delta U[t-1]$.
Equation~\eqref{eq:daq-app-matched-state} then gives
\begin{equation}
\Delta H[t-1]
=(1-S^{FP}[t-1])E[t-1]+\epsilon[t-1].
\label{eq:daq-app-inductive-state}
\end{equation}
Substitution into
Equation~\eqref{eq:daq-app-potential-error}
yields $\Delta U[t]=E[t]$.
Hence this equality holds whenever the preceding
firing histories agree.

If the outputs agree over all $T$ steps,
$E[t]=\Delta U[t]$ throughout the trajectory.
Since $U^Q[t]$ lies in the reference decision region,
\begin{equation}
U^{FP}[t]+E[t]\in\mathcal{D}_{S^{FP}[t]},
\qquad t=1,\ldots,T.
\label{eq:daq-app-trajectory-condition}
\end{equation}
This proves necessity.

Conversely, assume
Equation~\eqref{eq:daq-app-trajectory-condition}
holds at every step.
At $t=1$, $E[1]=\Delta U[1]$, so the region condition
ensures agreement.
Agreement through step $t-1$ gives $E[t]=\Delta U[t]$
by the preceding induction, and the region condition
then ensures agreement at step $t$.
This proves sufficiency.

For stateless I-LIF, $E[t]=\Delta U[t]$ at every evaluation,
so the result follows directly from
Equation~\eqref{eq:daq-region}.

In particular, if the first firing mismatch occurs
at step $\tau$, then $E[\tau]=\Delta U[\tau]$ because
the preceding histories agree.
The first violation of the region condition therefore
identifies the first firing mismatch.

\subsection{Firing-Error Bound}
\label{sec:daq-app-proposition-two}

Using the boundary weights and signed distances defined
in Sec.~\ref{sec:daq-objective}, let
\begin{equation}
\chi_{ej}=
\left|
\mathbf{1}[q_e\ge j]-\mathbf{1}[y_e\ge j]
\right|
\label{eq:daq-app-boundary-disagreement}
\end{equation}
indicate disagreement across boundary $j$.
Then
\begin{equation}
(q_e-y_e)^2
=
\sum_{j\in\mathcal{J}_e}
\omega_j(y_e)\chi_{ej}.
\label{eq:daq-app-ordinal-identity}
\end{equation}
If $q_e>y_e$, only boundaries
$j=y_e+1,\ldots,q_e$ contribute, with weights
$1,3,\ldots,2(q_e-y_e)-1$.
Their sum is $(q_e-y_e)^2$.
If $q_e<y_e$, the same odd integers appear in reverse
order over $j=q_e+1,\ldots,y_e$.
When $q_e=y_e$, all terms vanish.
Binary firing is the single-boundary case with unit weight.

\paragraph{Proof of Proposition 2.}
At a disagreeing boundary, either $y_e\ge j$ and $q_e<j$,
or $y_e<j$ and $q_e\ge j$.
The first case requires $x_e^Q\le b_j$,
and the second requires $x_e^Q\ge b_j$.

Since $\rho=\eta>0$, each disagreeing boundary satisfies
\begin{equation}
H_+\!\left(\frac{\rho-d_{ej}}{\eta}\right)
\ge H_+(1)=\frac12.
\label{eq:daq-app-mismatch-penalty}
\end{equation}
All remaining boundary terms are nonnegative.
Combining this observation with
Equation~\eqref{eq:daq-app-ordinal-identity} gives
\begin{equation}
\begin{split}
\frac{(q_e-y_e)^2}{\mu_e+\delta}
&=\frac{\sum_{j\in\mathcal{J}_e}
\omega_j(y_e)\chi_{ej}}{\mu_e+\delta}\\
&\le
\frac{2}{\mu_e+\delta}
\sum_{j\in\mathcal{J}_e}
\omega_j(y_e)
H_+\!\left(\frac{\rho-d_{ej}}{\eta}\right)\\
&=2\ell_e.
\end{split}
\label{eq:daq-app-event-bound-proof}
\end{equation}
Averaging over the same events and groups as
Equation~\eqref{eq:daq-aggregation} yields $R_w\le2L$.

\begin{table*}[t]
\centering
\caption{
Four-bit PTQ results on ImageNet-1K.
All entries report top-1 accuracy (\%).
W/A and W/U denote weight/activation and weight/membrane
bit widths, respectively.
ViT results follow the comparison in
FIMA-Q~\cite{wu2025fima}, with UQ-ViT results
from~\cite{jiang2026uq}.
All SDT-V3 models use $T\times D=1\times4$.
}
\label{tab:daq-vit-ptq}

\small
\setlength{\tabcolsep}{5pt}
\renewcommand{\arraystretch}{1.08}

\begin{tabular*}{\textwidth}{
    @{\extracolsep{\fill}}
    l c r r r r
    @{}
}
\toprule
\multicolumn{6}{c}{\textbf{ViT PTQ (W/A)}} \\
\midrule
Method
& Bits
& DeiT-T
& DeiT-S
& Swin-S
& ViT-B \\
\midrule

Full-Precision
& 32/32
& 72.71
& 79.85
& 83.23
& 84.54 \\
\midrule

PTQ4ViT~\cite{yuan2022ptq4vit}
& 4/4
& 36.96
& 34.08
& 76.09
& 30.69 \\

APQ-ViT~\cite{ding2023towards}
& 4/4
& 47.94
& 43.55
& 77.15
& 41.41 \\

RepQ-ViT~\cite{li2023repq}
& 4/4
& 57.43
& 69.03
& 79.45
& 68.48 \\

AdaLog~\cite{wu2024adalog}
& 4/4
& 63.52
& 72.06
& 80.77
& 79.68 \\

IGQ-ViT~\cite{moon2024instance}
& 4/4
& 62.45
& 74.66
& 80.98
& 79.32 \\

DopQ-ViT~\cite{yang2024dopq}
& 4/4
& 65.54
& 75.84
& 81.71
& 80.95 \\

UQ-ViT~\cite{jiang2026uq}
& 4/4
& 59.71
& 72.13
& 79.93
& 72.07 \\

FIMA-Q~\cite{wu2025fima}
& 4/4
& \textbf{66.84}
& \textbf{76.87}
& \textbf{81.82}
& \textbf{83.04} \\
\midrule

\multicolumn{6}{c}{\textbf{SDT-V3 PTQ (W/U)}} \\
\midrule
Method
& Bits
& SDT-V3-10M
& SDT-V3-19M
& SDT-V3-83M
& SDT-V3-171M \\
\midrule

Full-Precision
& 32/32
& 78.5
& 79.8
& 83.2
& 84.7 \\
\midrule

\textbf{DAQ (Ours)}
& 4/4
& 76.0
& 77.7
& 82.0
& 83.4 \\

\bottomrule
\end{tabular*}
\end{table*}

\section{Experimental Settings}
\label{sec:daq-experimental-settings}

\paragraph{Initialization.}
We initialize per-channel weight and membrane quantizers
by minimizing reconstruction error over percentile-based
candidates.
Membrane initialization uses all positions and time steps
of 128 calibration samples.

\paragraph{Sampling and optimization.}
We use 1,024 calibration samples and uniformly sample
up to eight positions per sample and neuron invocation,
shared across channels and time.
Calibration uses AdamW with zero weight decay and cosine
learning-rate decay for 320 updates at batch size 32.
VisEvent uses 1,280 updates at batch size 8.

\paragraph{Image classification.}
On ImageNet-1K, we evaluate pretrained
SDT-V3 models~\cite{yao2025scaling} with
$224\times224$ inputs and $T\times D=1\times4$.
The SDT-8-768 model used in the ablations takes
$288\times288$ inputs with $T\times D=4\times1$.
For CIFAR-10 and CIFAR-100, we use pretrained
SDT-2-512 models~\cite{yao2023spike} with LIF neurons
and $T=4$.
We follow each model's original preprocessing and
report top-1 accuracy on the standard validation
or test set.

\paragraph{Semantic segmentation.}
We evaluate SDT-V3-10M with an FPN neck on ADE20K,
using $T\times D=1\times4$.
W4/U4 quantization is applied to the backbone,
while the neck and segmentation head retain their
original precision.
We follow the original evaluation pipeline and
report mean intersection over union (mIoU) on
the validation set.

\paragraph{Event-based tracking.}
We evaluate SDTrack$_{\mathrm{Tiny}}$~\cite{shan2025sdtrack}
on FE108, FELT, and VisEvent, following its original
event representation and tracking protocol.
The template and search resolutions are
$128\times128$ and $256\times256$, respectively.
We consider LIF neurons with $T\times D=4\times1$
and I-LIF neurons with $T\times D=1\times4$.
Performance is measured by success AUC and precision
rate (PR) at a center-error threshold of 20 pixels.

\section{Comparison with ViT PTQ}
\label{sec:daq-vit-ptq}

Table~\ref{tab:daq-vit-ptq} summarizes four-bit PTQ
results on ImageNet-1K, with floating-point baselines
reported for each architecture.
The ViT methods quantize weights and activations (W4/A4),
while DAQ quantizes weights and membrane variables
at fixed bit widths (W4/U4).
PTQ4SNN~\cite{xie2026ptq4snn}, listed in
Table~\ref{tab:cifar_results}, combines 4-bit weights
with mixed-precision membrane quantization under an
average 4-bit budget, so comparisons with DAQ reflect
differences in both precision allocation and calibration.

Among the listed ViT methods, FIMA-Q achieves the highest
accuracy on all four architectures.
It obtains 81.82\% on Swin-S and 83.04\% on ViT-B,
corresponding to accuracy drops of 1.41 and 1.50
percentage points.
DAQ achieves 82.0\% on SDT-V3-83M and 83.4\% on
SDT-V3-171M, with drops of 1.2 and 1.3 points
from their respective floating-point baselines.
For the smaller SDT-V3-10M and SDT-V3-19M models,
DAQ obtains 76.0\% and 77.7\%, with drops of
2.5 and 2.1 points, respectively.

\section{Storage and Energy}
\label{sec:daq-resource-cost}

\subsection{Storage}
\label{sec:daq-storage-details}

We estimate the ideal-packed storage of model parameters
and membrane buffers at batch size one.
SDT-V3 uses $224\times224$ inputs with $T\times D=1\times4$,
while SDT-8-768 uses $288\times288$ inputs with
$T\times D=4\times1$.
Let $P_q$ denote the number of weights assigned $b_W$ bits
and $P_f$ the remaining FP32 parameters.

Membrane buffer capacity follows the neuron configuration.
For recurrent LIF, each node retains one post-reset state
for the next time step.
These buffers are reused across time steps, so their
capacity is independent of $T$.
For stateless I-LIF, we assume layerwise reuse of a
single buffer for pre-firing membrane potentials.
Its capacity is determined by the largest such tensor.

Let $\mathcal{R}$ and $\mathcal{I}$ denote the sets of
LIF and I-LIF nodes, respectively.
For node $\ell$, let $n_\ell$ be the number of elements
in its retained post-reset state for LIF, or its
pre-firing potential tensor for I-LIF.
Let $b_{U,\ell}$ denote the corresponding storage precision.
The storage costs are
\begin{equation}
\begin{aligned}
M_{\mathrm{param}}
&=b_WP_q+32P_f,\\
M_{\mathrm{mem}}
&=
\begin{cases}
\displaystyle
\sum_{\ell\in\mathcal{R}}b_{U,\ell}n_\ell,
& \text{LIF},\\[4pt]
\displaystyle
\max_{\ell\in\mathcal{I}}b_{U,\ell}n_\ell,
& \text{I-LIF},
\end{cases}\\
M_{\mathrm{total}}
&=M_{\mathrm{param}}+M_{\mathrm{mem}}.
\end{aligned}
\label{eq:daq-storage-cost}
\end{equation}
These quantities are measured in bits and converted
to MiB for reporting.
Other activations, operator workspaces, and quantizer
metadata are excluded.

$W_{\mathrm{FP}}$ denotes FP32 weights, and
$U_{\mathrm{FP}}$ follows the native
precision of each membrane tensor.
Under U4, each membrane element is assigned four bits.

\subsection{Energy Estimation}
\label{sec:daq-energy-details}

Following SDTrack and
TP-Spikformer~\cite{shan2025sdtrack,wei2026tp},
we estimate arithmetic energy from multiply--accumulate
(MAC) and accumulate (AC) operations.
We adopt the 45\,nm reference costs
$E_{\mathrm{MAC}}=4.6$\,pJ and
$E_{\mathrm{AC}}=0.9$\,pJ.

For a spike-driven convolutional or linear layer,
let $F_\ell$ be its dense operation count per time step
and $\bar{S}^{\mathrm{in}}_\ell$ its average presynaptic
firing output.
The synaptic operation count is
\begin{equation}
\mathrm{SOP}_\ell
=T\,\bar{S}^{\mathrm{in}}_\ell F_\ell.
\label{eq:daq-sop-count}
\end{equation}
For LIF, $\bar{S}^{\mathrm{in}}_\ell$ is the
average firing rate.
For I-LIF, it is the mean integer spike count before
output normalization.
We measure this activity in the quantized model.

Following the four-bit operation weighting in
QSD-Transformer~\cite{qiu2025quantized}, we estimate
the inference energy as
\begin{equation}
\begin{aligned}
E_{\mathrm{inf}}
&=E_{\mathrm{MAC}}N_{\mathrm{MAC}}\\
&\quad+E_{\mathrm{AC}}
\left(
\frac{N_{\mathrm{W4}}}{8}
+N_{\mathrm{FP}}
+N_{\mathrm{attn}}
\right).
\end{aligned}
\label{eq:daq-energy-cost}
\end{equation}
Here, $N_{\mathrm{W4}}$ and $N_{\mathrm{FP}}$ are
the raw synaptic operation counts for W4 and
unquantized convolutional or linear layers,
respectively.
$N_{\mathrm{MAC}}$ counts dense MAC operations,
and $N_{\mathrm{attn}}$ accounts for spike-driven
attention.
The $1/8$ factor applies only to the W4 synaptic
operations; dense operations and attention retain
their respective costs.

Operation counts follow the corresponding architecture's
energy model.
For SDT-V3, we use its logical projection structure and
the SDSA operation count, assuming W4 execution of the
logical projections.
Tracking follows SDTrack's temporal accounting.
Segmentation energy is evaluated at $512\times512$,
including the floating-point neck and prediction head.
We report energy per image or tracked frame.
Membrane quantization affects this estimate through
the measured firing activity.

\end{document}